\documentclass[10pt,twocolumn,letterpaper]{article}

\usepackage[T1]{fontenc}

 \usepackage[pagenumbers]{cvpr} 

\definecolor{cvprblue}{rgb}{0.21,0.49,0.74}
\usepackage[pagebackref,breaklinks,colorlinks,allcolors=cvprblue]{hyperref}

\def\paperID{*****} 
\def\confName{CVPR}
\def\confYear{2026}

\title{Results of the 1st Asynchronous CASTLE Challenge at the Joint Egocentric Vision Workshop in Conjunction with CVPR 2026}

\author{Luca Rossetto\\
Dublin City University\\
Dublin, Ireland\\
{\tt\small luca.rossetto@dcu.ie}
\and
Werner Bailer\\
JOANNEUM RESEARCH\\
Graz, Austria\\
{\tt\small werner.bailer@joanneum.at}
\and
Cathal Gurrin\\
Dublin City University\\
Dublin, Ireland\\
{\tt\small cathal.gurrin@dcu.ie}
\and
Graham Healy\\
Dublin City University\\
Dublin, Ireland\\
{\tt\small graham.healy@dcu.ie}
\and
Omar Shahbaz Khan\\
IT University of Copenhagen\\
Copenhagen, Denmark\\
{\tt\small omsh@itu.dk}
\and
Stevan Rudinac\\
University of Amsterdam\\
Amsterdam, The Netherlands\\
{\tt\small s.rudinac@uva.nl}
\and
Klaus Schöffmann\\
Klagenfurt University\\
Klagenfurt, Austria\\
{\tt\small klaus.schoeffmann@aau.at}
\and
Allie Tran\\
Dublin City University\\
Dublin, Ireland\\
{\tt\small allie.tran@dcu.ie}
}

\begin{document}
\maketitle

\begin{abstract}
This report summarizes the contributions and results of the 1st Asynchronous CASTLE Challenge at the Joint Egocentric Vision Workshop in conjunction with CVPR 2026.

\end{abstract}

\section{Introduction}

The CASTLE Challenge @ EgoVis 2026, held as a pilot task co-located with the EgoVis CVPR 2026 workshop\footnote{\url{https://egovis.github.io/cvpr26/}}, focused on closed-form Question Answering (QA) over long-term, multi-modal egocentric data. The challenge leveraged the CASTLE 2024 dataset, a comprehensive collection of high-fidelity, multi-perspective recordings captured in a controlled environment over four days. 

\subsection{Dataset}
The CASTLE 2024 dataset~\cite{DBLP:conf/mm/RossettoBDH0KLR25} consists of over 600 hours of Ultra-High Definition (UHD) video recorded at 50 frames per second, accompanied by synchronized audio. The data captures 10 participants engaged in natural, everyday activities. The recording setup utilized 15 distinct video streams: 10 from egocentric cameras (worn by the participants) and 5 from static perspective cameras. To further enrich the multi-modal experience, the dataset is augmented with 6DoF IMU data, GPS coordinates, biometric signals (e.g., heart rate), gaze data, and thermal imagery, providing a robust testbed for comprehensive scene and activity understanding.

\subsection{Task Definition}
The pilot task challenged participants to solve a closed-form Question Answering problem. The dataset was accompanied by a benchmark of 185 multiple-choice questions (4 choices), each requiring systems to reason across the multi-modal, multi-view inputs to identify the correct answer from four provided options. The questions were designed to test various capabilities, including temporal reasoning, object tracking, visual counting, and spatial awareness. For example, participants had to answer queries such as:

\begin{itemize}
    \item ``\textit{At what rate was Cathal charging his car on the last day?}'' -- testing numerical and temporal reasoning.
    \item ``\textit{Who had the first slice of Christmas cake?}'' -- testing activity recognition and participant identification.
    \item ``\textit{Where was the little blue whale plushie hiding on the second day?}'' -- testing object detection and spatial memory across extended timeframes.
\end{itemize}
     
Accuracy is used to evaluate the performance of each submission.
     
\section{Team Contributions}

The challenge received significant community interest, attracting 43 registered participants who generated a total of 272 submission entries on the leaderboard. By the end of the challenge, four teams successfully submitted technical reports detailing their methodologies. 

The final leaderboard was highly competitive. The top-performing teams achieved scores between 50\% and 58\% accuracy, demonstrating that while current systems are capable of navigating 600 hours of multi-view data, long-context, closed-form QA in unscripted environments remains a highly challenging problem. Below, we summarize the approaches and primary contributions of the four reporting teams, ranked by their final leaderboard performance.

\subsection{Team WDL}

Team WDL~\cite{DBLP:journals/corr/abs-2606-00712} proposed an evidence-aware multimodal reasoning pipeline centered on the Qwen multimodal large language model. Their core contribution lies in transforming the unstructured long-form QA task into a structured evidence retrieval and verification problem using an official-data-only pipeline.

\paragraph{Evidence-Aware Context Construction} The system parses explicit and implicit hints from the query (e.g., person names, days, rooms, visual/speech keywords) to guide retrieval. Official ASR transcripts are segmented and scored based on lexical overlap, phrase-level matches, and contextual bonuses. Auxiliary photos are ranked using path-level hints, while video frames are uniformly sampled from matched candidate videos.
     
\paragraph{Question-Type Routing} The team introduced a specialized routing mechanism that categorizes queries into static visual, speech/text, temporal, or mixed types. Each type triggers a tailored instruction block (e.g., temporal prompts reconstruct order from timestamps, while static visual prompts prioritize object counts and OCR).
     
\paragraph{LoRA Adaptation and Self-Consistency} To align the base model with the CASTLE multiple-choice format, the team employed LoRA (Low-Rank Adaptation) fine-tuning, computing loss only on answer tokens. They also utilized multi-sample self-consistency, running strict, balanced, and aggressive prompt variants for each question. The final prediction was aggregated via confidence-weighted voting. Ablation studies confirmed that LoRA adaptation improved accuracy from 21\% to 50\%, and increasing the count of sampled video frames further increased performance to 58\%, highlighting evidence coverage as the main bottleneck.

\subsection{Team MARS} 

Team MARS (Multimodal Agentic Reasoning with Source selection) \cite{DBLP:journals/corr/abs-2605-18176} primary contribution is a source-selective agentic framework that dynamically decides which evidence modality to query next, rather than processing all raw media simultaneously.

\paragraph{Source Taxonomy and Memory Construction} MARS organizes data into two primary sources (video summaries and transcripts) and four auxiliary sources (gaze, heart rate, photos, and thermal imagery). Long videos are processed offline using the HCQA pipeline -- sliced into clips, captioned, and compressed into event summaries using DeepSeek. Transcripts are normalized into utterance windows with preserved timestamps.
     
\paragraph{Agentic Decision Loop} At inference time, a GPT-5.4 decision agent operates in an iterative loop, selecting from four actions: Continue thinking (refine the retrieval query without loading new data), Add data (request a specific modality, such as thermal imagery or gaze, for a target time/person), Answer (select an option and cite supporting evidence), or Random fallback (for unsupported cases).
     
\paragraph{Solution Evolution Insights} The team provided a detailed evolution of the accuracy of their system. Starting from a baseline of 35\% text-only, the addition of official transcripts yielded a gain of 7\%. A further 4\% was achieved by parsing the cues of the day, person, and room to narrow the retrieval. However, the final jump from 55\% to 57\% was unlocked by the full MARS agentic loop, proving that dynamically requesting missing modalities (like heart rate for physical exertion questions or gaze for attention queries) is highly effective for complex, multi-modal reasoning.

\subsection{Team TAHAKOM}

Team TAHAKOM~\cite{DBLP:journals/corr/abs-2606-01933} developed a training-free agentic framework whose main contribution is a hierarchical knowledge graph (KG) retrieval mechanism designed to map static and dynamic entities across long temporal contexts.

\paragraph{Multimodal Knowledge Graph Construction} The team omitted exocentric streams, relying solely on egocentric video divided into 30-second clips. A multimodal extraction function jointly processed visual and audio modalities to generate clip-level captions. These were then structured into a directed KG stored in an SQLite database. The graph maps entity types (person, location, object, food, activity event, action) and relationship types (talks-to, interacts-with, mentions, uses, located-in) with strict temporal intervals.
     
\paragraph{Hierarchical Agentic Workflow} Built on LangGraph, their agentic framework decomposes queries into sub-tasks via a PlannerAgent. A GraphRetrieval agent then executes a ``strict-to-relaxed'' SQL generation strategy: it first attempts exact database field matches, gradually relaxing constraints if no records are returned. 
     
\paragraph{Ablation and Structural Findings} The team conducted rigorous examinations of their KG structure. They found that including ``Mention'' relationships in the database improved accuracy by 7\%, as it provided critical structural data to pinpoint specific speakers and temporal intervals that aggregated transcripts missed. Furthermore, adding a ClipSearch and ClipAnalyze module -- which uses the KG's temporal boundaries to extract and evaluate raw video clips for visual queries like counting and OCR -- yielded a 4\% performance boost, validating the necessity of falling back to raw pixels for visually grounded tasks.

\subsection{Team CuriosAI} 

Team CuriosAI~\cite{DBLP:journals/corr/abs-2605-27800} explored two distinct approaches built atop a shared, five-lane multi-modal preprocessing layer (person identity, visual captions, detected objects, audio transcripts, and action timelines). Their primary contribution is an explicit, multi-stage pipeline designed to combat the confabulation tendencies of large VLMs in long-video contexts.

\paragraph{SVA (Search–Verify–Answer) Pipeline} Their final submission (SVA) decouples the QA process into three explicit stages. Search aggregates preprocessing into 50 (day, hour) cells, using BM25 and e5-large-v2 hybrid retrieval, reranked by GPT-5-mini to find a primary 15-minute window. Verify expands this anchor into $\sim$ 24 adjacent 5-minute sub-windows across cameras. Answer fuses the evidence using a GPT-5 judge following an evidence-priority hierarchy (OCR $\rightarrow$ audio quote $\rightarrow$ visual $\rightarrow$ context).
     
\paragraph{Anti-Confabulation Discipline} A major innovation of their work is the prompt-level rules enforced during the Verify stage. Unconstrained VLMs frequently failed by re-quoting prompt context, asserting choices on silent/test-pattern clips, or counting without spatial grounding. To fix this, the team enforced four rules: no echo (do not re-quote prompt context), abstain (do not answer on silent clips), localize (require spatial grounding for counts), and ground (ungrounded high confidence is rejected).
     
\paragraph{Pipeline Contrast (SVA vs. TMKG)} The team contrasted SVA with a secondary approach, TMKG (Temporal–Multimodal–Knowledge–Graph), which packages observations into a temporal knowledge graph and answers with a single grounded VLM. SVA achieved 50\% accuracy (requiring $\sim 28$ LLM/VLM calls per question) compared to TMKG's 35\% ($\sim 1$ call per question). This 15\% gap demonstrated that disciplined, multi-step clip-level verification is a layer worth adding on top of any retrieval substrate, despite its higher computational cost.

\section{Result Analysis}

The official leader-board is shown in Table \ref{tab:results}.
The ranked teams were able to answer between 65 and 108 of the 185 questions correctly, resulting in an accuracy ranging from 0.351 to 0.584.
All teams therefore substantially outperformed the random baseline of 0.25.

\begin{table}[]
\centering
\caption{Official Challenge Results}
\label{tab:results}
\begin{tabular}{|l|r|r|}
\hline
Team     & \multicolumn{1}{l|}{Correct Answers} & \multicolumn{1}{l|}{Accuracy} \\ \hline
WDL & 108                                  & 0.584                         \\ \hline
MARS     & 105                                  & 0.568                         \\ \hline
TAHAKOM  & 101                                  & 0.546                         \\ \hline
CuriosAI (SVA) & 92                                   & 0.497                         \\ \hline
CuriosAI (TMKG) & 65                                   & 0.351                         \\ \hline
\end{tabular}
\end{table}

The overlap analysis between the highest scoring submissions shows clear differences in the approaches.
Although the highest scoring team managed to answer 108 questions correctly, the teams in aggregate could correctly solve 176 questions.
This means that there were only 9 questions that were not correctly answered by any of the ranked teams.
Table \ref{tab:team-aggregate} shows an overview of how many questions were correctly answered by how many teams.

\begin{table}[]
\centering
\caption{Aggregate histogram over correctly answered questions and teams.}
\label{tab:team-aggregate}
\begin{tabular}{|r|r|}
\hline
Teams answered correctly & Number of Questions \\ \hline
0                        & 9                   \\ \hline
1                        & 34                  \\ \hline
2                        & 48                  \\ \hline
3                        & 47                  \\ \hline
4                        & 35                  \\ \hline
5                        & 12                  \\ \hline
\end{tabular}
\end{table}

\section{Conclusion}

In this report, we provided an overview of the first CASTLE question answering challenge, held at the joint EgoVis workshop at CVPR 2026.
The presented result analysis suggests that, while the submitted submissions received were already promising, a combination of the evaluated approaches might be even more effective.
We are looking forward to evaluating this and other ideas in future editions of the challenge.

\clearpage

{
    \small
    \bibliographystyle{ieeenat_fullname}
    \bibliography{main}
}

\end{document}